\documentclass[mathematics,article,accept,pdftex,moreauthors]{mdpi} 
\firstpage{1} 
\pubvolume{1}
\issuenum{1}
\articlenumber{0}
\pubyear{2022}
\copyrightyear{2022}
\externaleditor{ } 
\datereceived{} 
\dateaccepted{} 
\datepublished{} 
\hreflink{} 

\usepackage{xcolor}
\usepackage{bm}
\usepackage{algorithmic}
\usepackage[ruled,vlined]{algorithm2e}
\usepackage{soul}
\Title{Dendrite Net with Acceleration Module for Faster Nonlinear Mapping and System Identification}

\TitleCitation{Dendrite Net with Acceleration Module for Faster Nonlinear Mapping and System Identification}

\Author{Gang Liu 
 $^{1,2}$\orcidA{}, Yajing Pang $^{1}$, Shuai Yin $^{2}$, Xiaoke Niu $^{1,3,}$, Jing Wang $^{2,}$ and Hong Wan $^{1,3,}$}

\AuthorNames{Gang Liu, Yajing Pang, Shuai Yin, Xiaoke Niu, Jing Wang and Hong Wan}

\AuthorCitation{Liu, G.; Pang, Y.; Yin, S.; Niu, X.; Wang, J.; Wan, H.}

\address{%
$^{1}$ \quad School of Electrical and Information Engineering, Zhengzhou University, Zhengzhou 450001, China; 
\\
$^{2}$ \quad Institute of Robotics and Intelligent Systems, Xi’an Jiaotong University, Xi'an 710049, China;\\
$^{3}$ \quad Henan Provincial Key Laboratory of Brain Science and Brain-Computer Interface Technology, Zhengzhou~450001, China }

\abstract{Nonlinear mapping is an essential and common demand in online systems, such as sensor systems and mobile phones. Accelerating nonlinear mapping will directly speed up online systems. Previously the authors of this paper proposed a  Dendrite  Net (DD) with enormously lower time complexity than the existing nonlinear mapping algorithms; however, there still are redundant calculations in DD. This paper presents a DD with an acceleration module (AC) to accelerate nonlinear mapping further. {{We conduct three experiments to verify whether DD with AC has lower time complexity while retaining DD's nonlinear mapping properties and system identification properties}}: The first experiment is the precision and identification of unary nonlinear mapping, reflecting the calculation performance using DD with AC for basic functions in online systems. The second experiment is the mapping precision and identification of the multi-input nonlinear system, reflecting the performance for designing online systems via DD with AC. Finally, this paper compares the time complexity of DD and DD with AC and analyzes the theoretical reasons through repeated experiments. \textit{Results:} 
 DD with AC retains DD's excellent mapping and identification properties and has lower time complexity.  \textit{Significance:} DD with AC can be used for most engineering systems, such as sensor systems, and will speed up computation in these online systems. The code of DD with AC is available on 
\textit{GitHub:\href{https://github.com/liugang1234567/Gang-neuron}{Gang neuron}}.}


\begin{document}

\section{Introduction}

The development of online systems, such as sensor systems, mobile phones, and computers, is changing the world we live in~\cite{5306098,5560598}. Nowadays, attention has been paid to the running speed of online systems, as  running speed is the evaluation index for system performance. Nonlinear mapping is an essential and common demand in online systems, such as basic function calculation (e.g., $y=sin(x) $, where map $ x $ to $ y $) in computers~\cite{sammon1969nonlinear,DCOM11,penJDK}. The time complexity of nonlinear mapping directly affects the running speed of online systems~\cite{lim2000comparison}.  However,  after years of development, the speed of nonlinear mapping is becoming increasingly difficult to improve, and there is less and less research on improving the running speed of online systems by accelerating the nonlinear mapping speed.

{It is typical to store and calculate basic nonlinear functions (nonlinear mapping) in a polynomial form in computers (e.g., $ sin(x) $ in C} ~\cite{DCOM11}{, or $ sin(x) $ in java} ~\cite{penJDK}). This means that polynomial storage and calculation methods with lower time complexity will speed up the running speed of online systems.  In mathematics and computer science, Horner's method (or Horner's scheme) is a well-established algorithm with lower time complexity ($ O(n) $) for polynomial evaluation~\cite{cajori1911horner}. {In 2021, the authors of this paper proposed DD}~\cite{liu2020dendrite}. DD can be seen as a polynomial form with lower time complexity than the traditional polynomial form, and the time complexity of DD is consistent with Horner's method. During the year, DD has already been used in multiple areas, such as energy saving~\cite{Han2022}, spatiotemporal traffic flow data imputation~\cite{wang2022hybrid}, high-dimensional problems~\cite{zhang2022adaptive}, image processing~\cite{li2021semantic}, multi-objective optimization~\cite{ding2022multi}, accuracy prediction of the RV reducer to be assembled~\cite{Jin2022}, and precipitation correction in flood season~\cite{li2022algorithm}. Nevertheless, there are still some redundant calculations in DD when the order of the DD model is higher than the number of inputs.

\textls[-15]{This paper proposes an acceleration module for DD to reduce redundant calculation, named DD with AC, which further speeds up the computation in online systems. According to the theory of DD, DD with AC can be used for nonlinear mapping and system identification.
Meanwhile, the time complexity of DD with AC should be lower than DD according to the aim of this study. Consequently, the corresponding theory of DD with AC is explored by the experiments here. The main contributions of this paper are presented as follows:}
\begin{enumerate}
\item {In this paper, the redundant calculations in DD are found,} and some characteristics of the redundant calculations are given;

\item This paper {presents} an acceleration module for DD to reduce redundant calculations and presents DD with AC by theoretically analyzing the redundant calculations in DD;

\item {The proposed concept is experimentally justified and computationally verified based on the theoretical analysis. After theoretical and experimental analysis, it is demonstrated that DD with AC can be used for nonlinear mapping and system identification with lower time complexity than DD.}
\end{enumerate}

The rest of the paper is organized as follows. Section~\ref{sec2} introduces DD and describes the design of DD with AC. Experiments and results are given in Section~\ref{sec3}. Section~\ref{sec4}  discusses some experimental results and the significance of this study, and the conclusions are drawn in Section~\ref{sec5}.

\section{{Design} of Dendrite Net with Acceleration Module} \label{sec2}

\subsection{Dendrite Net and Its Redundant Calculations}

\subsubsection{Dendrite Net}

{In a previous study, the authors of this paper proposed a basic machine learning algorithm called Dendrite Net or DD with white-box properties, controlled accuracy to improve generalization, and low computational complexity}~\cite{liu2020dendrite}. DD's main concept is that if the output's logical expression contains the logical relationship of a class among inputs (\textit{and$\backslash$or$\backslash$not}), the algorithm can recognize the class after learning~\cite{liu2020dendrite}. DD is one of the dendrites of a Gang neuron (an improved artificial neuron)~\cite{liu2020may}, {and its essence is a specialized polynomial form.}

\begin{figure}[H]
	\includegraphics[width=0.8\columnwidth]{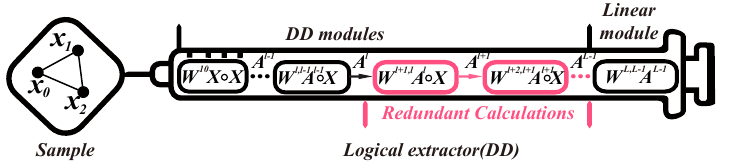}
	\caption{Schematic of dendrite net. DD aims to design the output's logical expression of the corresponding class (logical extractor). }
	\label{DD}
\end{figure}

{DD consists of DD modules and linear modules} (see Figure~\ref{DD}). The DD module is straightforward and is expressed as follows.

\begin{equation} 
	\bm{A}^{l}=\bm{W}^{l,l-1}\bm{A}^{l-1} \circ \bm{X}
	\label{DDmo}
\end{equation}
where $ \bm{A}^{l-1} $ and $ \bm{A}^{l} $ are the inputs and outputs of the module. $ \bm{X} $ denotes the inputs of DD. One of the elements in $ \bm{X} $ can be set to 1 to generate a bias.
$ \bm{W}^{l,l-1} $ is the weight matrix from the  $ (l-1)$-th module to the $ l$-th module.” $ \circ $ denotes the Hadamard product. Hadamard product is used to construct interactive items (e.g., $ x_{1}x_{2} $). The last module of DD is the linear module, {and }the linear module is expressed as follows.

\begin{equation} 
	\bm{A}^{L}=\bm{W}^{L-1,L}\bm{A}^{L-1}
	\label{linmo}
\end{equation}
where $ \bm{A}^{L-1} $ and $ \bm{A}^{L} $ are the inputs and outputs of the module. $ L $ expresses the number of modules. $ \bm{W}^{L,L-1} $ is the weight matrix from the  $ (L-1)$-th module to the $ L$-th module.

The following set of equations describes the gradient descent rule of DD.

\textit{The forward propagation of DD module and linear module:} 

\begin{equation} 
	\left\{
	\begin{aligned}
		\bm{A}^{l}&=\bm{W}^{l,l-1}\bm{A}^{l-1} \circ \bm{X} \\ 
		\bm{A}^{L}&=\bm{W}^{L,L-1}\bm{A}^{L-1}
	\end{aligned}
	\right.
	\label{DDeq.3}
\end{equation}

\textit{The error-backpropagation of DD module and linear module:}

\begin{equation} 
	\bm{dA}^{L}=\bm{\widehat{Y}}-\bm{Y}
	\label{DDeq.4}
\end{equation}
\begin{equation} 
	\left\{
	\begin{aligned}
		\bm{dZ}^{L}&=\bm{dA}^{L} \\ 
		\bm{dZ}^{l}&=\bm{dA}^{l} \circ \bm{X} 
	\end{aligned}
	\right.
	\label{DDeq.5}
\end{equation}
\begin{equation} 
	\bm{dA}^{l-1}=(\bm{W}^{l,l-1})^{T}\bm{dZ}^{l}
	\label{DDeq.6}
\end{equation}

\textit{The weight adjustment of DD:}

\begin{equation} 
	\bm{dW}^{l,l-1}=\frac{1}{m}\bm{dZ}^{l}(\bm{A}^{l-1})^{T}
	\label{DDeq.7}
\end{equation}
\begin{equation} 
	\bm{W}^{l,l-1(new)}=\bm{W}^{l,l-1(old)}-\alpha \bm{dW}^{l,l-1}
	\label{DDeq.8}
\end{equation}
{where} \bm{$ \widehat{Y} $} and \bm{$ Y $} are DD’s outputs and labels, respectively. $ m $ denotes the number of training samples in one batch. The learning rate $ \alpha $ can either be adapted with epochs or fixed to a small number based on heuristics. 

\begin{figure}[H]
	\includegraphics[width=0.8\columnwidth]{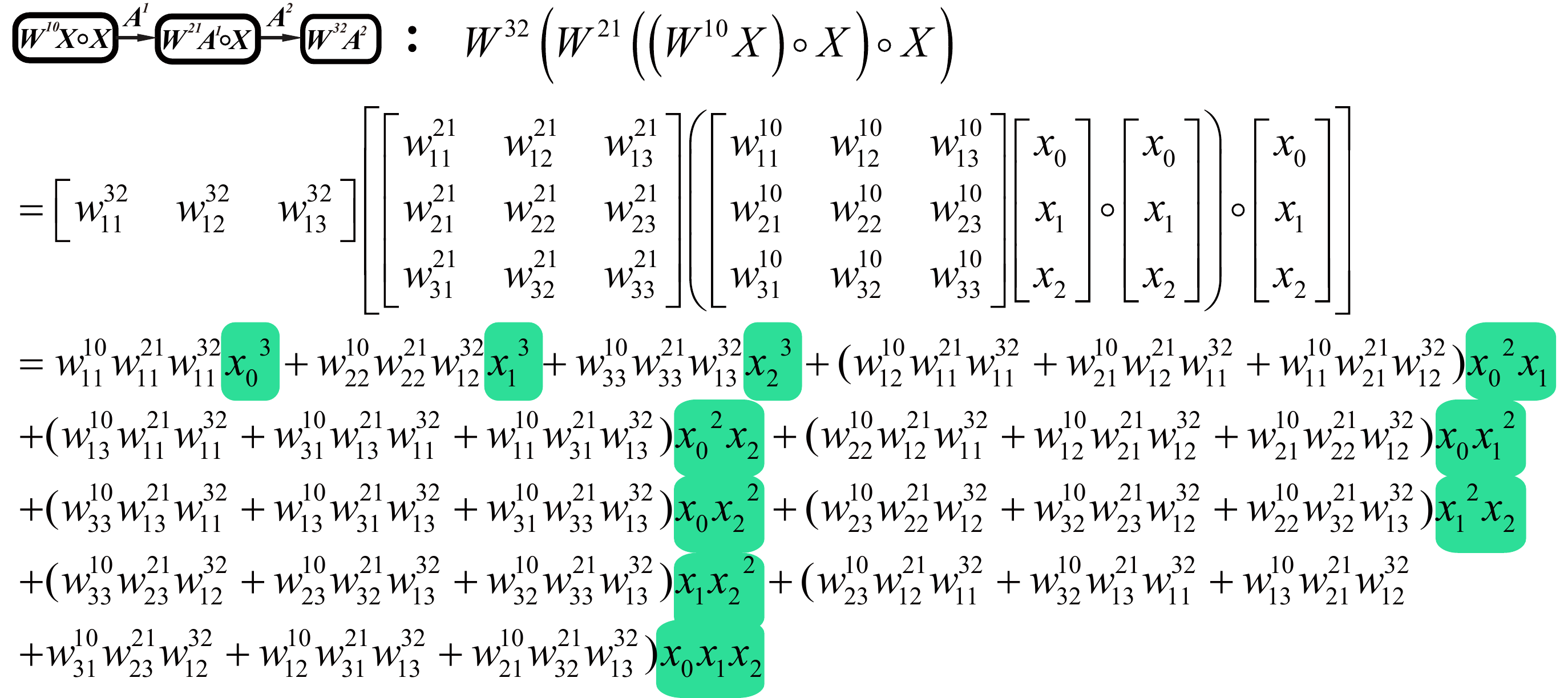}
	\caption{The   
 expanded form of DD.}
	\label{DDFig2}
\end{figure}

\begin{figure}[H]
	\includegraphics[width=0.8\columnwidth]{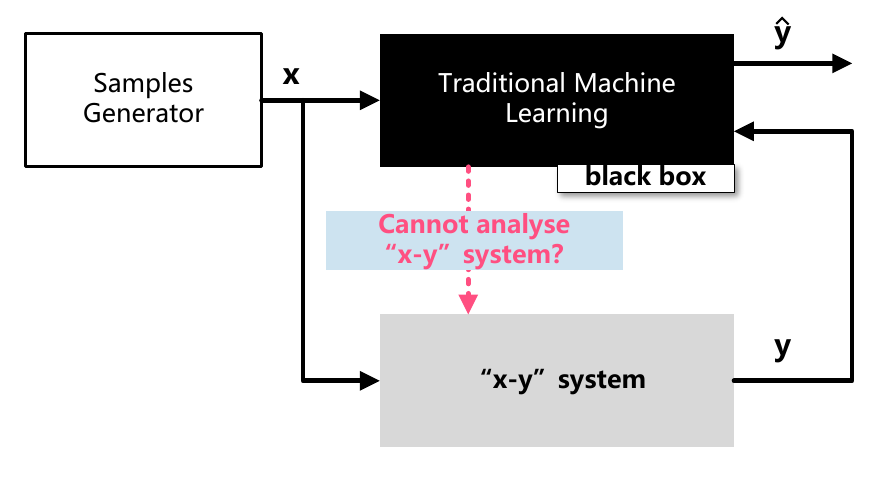}
	\caption{“Black-box” issue of traditional machine learning (ML)~\cite{rudin2019stop}. Traditional ML can generate $ \hat{y} $ approaching $ y $ via $ x $, but cannot analyze the ``$ x$-$y $'' system. Interestingly, the trained DD model can be transformed into a Relation spectrum such as the Taylor series for system identification~\cite{liu2020relation,Liu2021}. [{Analogous to the Fourier transform and Fourier spectrum used to decompose the signal, the DD and the Relation spectrum decompose the system.}].
	}
	\label{Fig1}
\end{figure}

The most attractive feature is that there are only matrix multiplication and Hadamard product in DD operation, which confers a white-box property and lower time complexity onto DD. \textit{White-box property}: The trained DD model can be translated into the Relation spectrum about inputs and outputs by formula simplification with software (e.g., \textit{MATLAB}, an example in Figure \ref{DDFig2}).  Concretely, the optimized weights are assigned to the corresponding matrices in Equations~(\ref{DDmo}) and (\ref{linmo}). Then the Relation spectrum was obtained through formula simplification in software. The white-box property solves the “black-box” issue of ML in Figure~\ref{Fig1}~\cite{rudin2019stop}; thus, DD integrates nonlinear mapping/pattern recognition and system identification, compared to other ML algorithms (see Table~\ref{Tableb}). {The lower time complexity makes it possible for DD to become a common algorithm in online systems.}

\unskip

\unskip

\begin{table}[H] 
	\caption{Comparison between DD and typical algorithms.\label{Tableb}}
	\newcolumntype{C}{>{\centering\arraybackslash}X}
	\begin{tabularx}{\textwidth}{CCCC}
		\toprule
		\textbf{Algorithms}	& \textbf{Readability}	& \textbf{Nonlinear Mapping (Online)} & \textbf{Analysis (Offline)}\\
		\midrule
		SVM, traditional NN, etc.~\cite{rudin2019stop,noble2006support} & Black box & Yes & No   \\
		Fourier transform and Fourier Spectrum~\cite{bracewell1986fourier} & White box & No & Yes (decomposing signal)\\
		DD and Relation spectrum~\cite{liu2020relation,liu2020dendrite,Liu2021,Han2022,wang2022hybrid,zhang2022adaptive,li2021semantic,ding2022multi,Jin2022,li2022algorithm} & White box & Yes & Yes (decomposing system$\backslash$model)\\
		\bottomrule
	\end{tabularx}
\end{table}

\subsubsection{Redundant Calculations in Dendrite Net}

\begin{figure}[H]
\includegraphics[width=0.9\columnwidth]{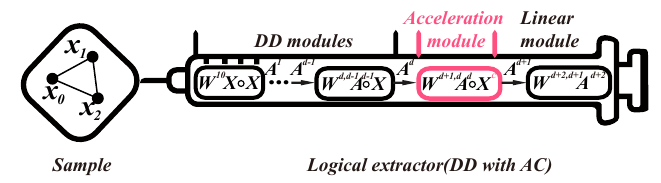}
	\caption{Schematic of the dendrite net with the acceleration module. }
	\label{Fig2}
\end{figure}

{DD constructs the interaction terms of the input variables and increases the order by concatenating DD modules} (see Figures~\ref{DD} and \ref{DDFig2})~\cite{liu2020dendrite}. {However, when the order of the DD model is higher than the number of inputs, the previous DD module will build all the interaction terms, while the later DD module will only increase the order without increasing the interaction terms.} For instance, in Figure \ref{DD}, if the number of inputs is $ l+1 $, $ l $ DD modules have constructed all the interactive items (the order of $ l $ DD modules is $ l+1 $.), the later red DD modules will only increase the order and cannot increase the interactive items. Hence, the later DD modules have redundant calculations, and we can increase the order with lower computation than the DD module. { Therefore, we replace the later DD modules with an acceleration module to increase the order faster in this paper} (see Figure~\ref{Fig2}).

\subsection{Dendrite Net with Acceleration Module}

\subsubsection{Architecture}

\begin{figure}[H]
\includegraphics[width=0.88\columnwidth]{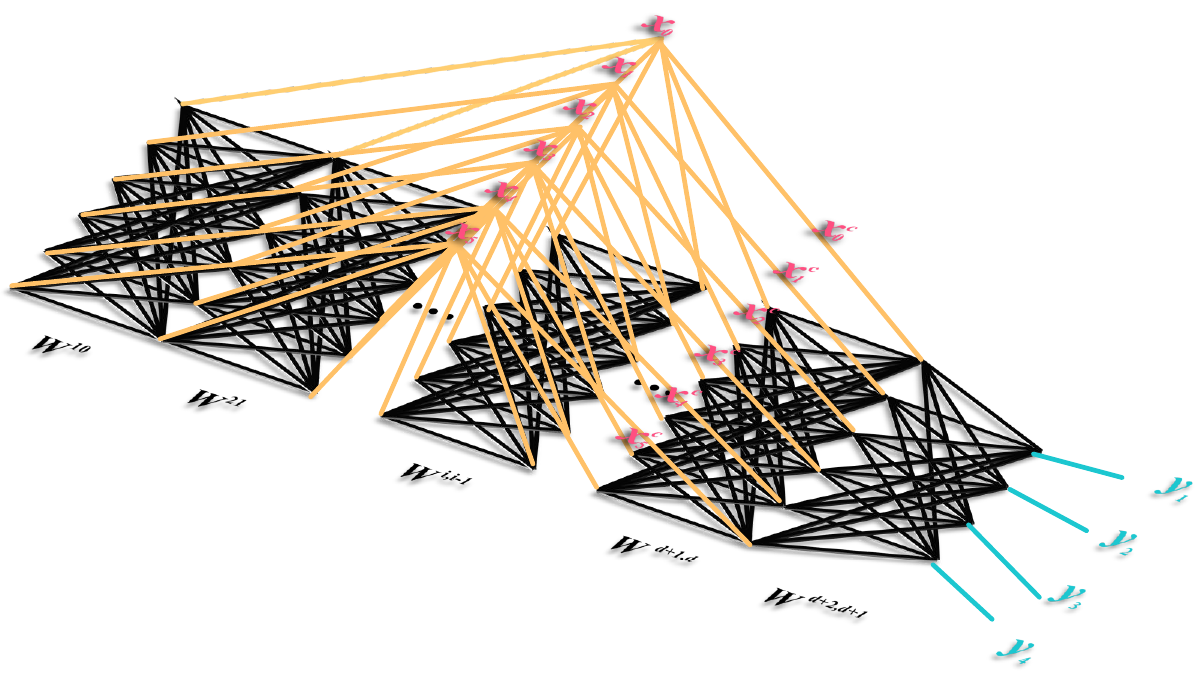}
	\caption{Overall architecture of DD with AC with six inputs and four outputs. {This figure is a visual example of Equation}~\eqref{eshu}.
	}
	\label{shu}
\end{figure}

Figure~\ref{Fig2} shows the dendrite net with the acceleration module. It contains DD modules, a linear module, and an acceleration module. The overall architecture of DD with the acceleration module is shown in Figure \ref{shu}. The architecture can be represented according to the following formula.

\begin{equation} 
		\label{eshu}
		\bm{Y}=\bm{W}^{d+2,d+1}[\bm{W}^{d+1,d}(\cdots \bm{W}^{21}(\bm{W}^{10}\bm{X} \circ \bm{X}) \circ \bm{X} \cdots) \circ \bm{X}^{c} ], d\in N+
\end{equation}
where $ \bm{X} $ and $ \bm{Y} $ denote the input space and the output space. One of the elements in $ \bm{X} $ can be set to 1 to generate a bias. $ \bm{W}^{i,i-1} $ is the weight matrix from the  $ (i-1)$-th module to the $ i$-th module. The last module is linear.   $d $ denotes the number of DD modules. $ c $ denotes Power of Number.  $ \circ $ denotes Hadamard product. 

The DD modules and the linear module {have} been described previously. {The acceleration module is expressed as follows:}

\begin{equation} 
	\bm{A}^{d+1}=\bm{W}^{d+1,d}\bm{A}^{d} \circ \bm{X}^{c}
	\label{eq.1}
\end{equation}
where $ \bm{A}^{d} $ and $\bm{A}^{d+1} $ are the inputs and outputs of the module. $ \bm{X} $ denotes the inputs of DD with AC. One of the elements in $ \bm{X} $ can be set to 1 to generate a bias. $d $ denotes the number of DD modules. $ c $ denotes Power of Number.  $ \bm{X}^{c} $ represents $ \bm{X} $ to the power of $ c $. $ \bm{W}^{d+1,d} $ is the weight matrix from the  $ d$-th module to the $d+1$-th module. $ \circ $ denotes Hadamard product.

{In order for the DD with AC to include all terms under the target order and use fewer modules,} the order of DD with AC $ n $, the number of DD modules $ d $, the power of AC $ c $, and the input dimension of DD with AC $ a $ should satisfy the following equations:

\begin{equation} 	
	\left\{
	\begin{aligned}
		&d+1+c=n\\
		&(c-1)\times a<n\\
		&c\times a\ge n
	\end{aligned}
	\right.
	\label{eq.2}
\end{equation}
where $ d+1+c=n $ means that the order of DD with AC is equal to the sum of the order of DD modules ($ d+1 $) and the order of the acceleration module ($ c $). $ (c-1)\times a<n $ means that if $ c $ exceeds this range (the value of $ c $ is too large), DD with AC can not construct all the interactive items. $ c\times a\ge n $ means that if the value of $ c $ is too small, there are still redundant computations. $ d $ and $ c $ are calculated by {Algorithm~\ref{algorithm1}} in the case of a given order $ n $.
\vspace{12pt} 

\begin{algorithm}[H]
	\caption{Design of DD with AC}\label{algorithm1}
	\LinesNumbered 
	\KwIn{The order of DD with AC $ n $, the input dimension of DD with AC $ a $}
	\KwOut{The number of DD modules $ d $, the power of AC $ c $}
	\For{i = 0 to $ n $}{
		\If{$(i-1)\times a<n$ and $i\times a\ge n$} {
			$c \gets i$\\
			$ d \gets n-c-1 $\\
			\If{$d<1$} {
				$d \gets 1$\\
				$ c \gets n-d-1 $\\	
			}	
		}	
		
	}
	\label{a1}
\end{algorithm}

\subsubsection{Learning Rules}

\begin{figure}[H]
\includegraphics[width=\columnwidth]{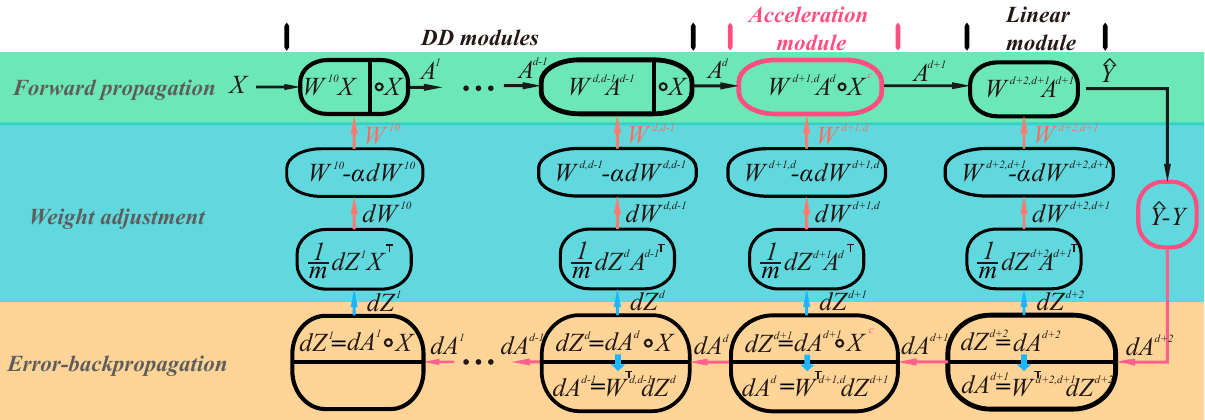}
	\caption{Graphical illustration of the learning rule.} 
	\label{DCFig5}
\end{figure}

The graphical illustration of the learning rule is shown in Figure~\ref{DCFig5}. As an example, we use {one-half} of the mean squared error (MSE) as the loss function. The learning rules of DD modules and {the} linear module have been described previously. The following set of equations describes the error back-propagation-based learning rule of the acceleration module (see Figure~\ref{DCFig5})~\cite{lecun1988theoretical}.

\textit{The error back-propagation of the acceleration module:}  

\begin{equation} 
	d\bm{Z}^{d+1}=d\bm{A}^{d+1} \circ \bm{X}^{c} 
	\label{eq.3}
\end{equation}
\begin{equation} 
	d\bm{A}^{d}=(\bm{W}^{d+1,d})^{T}d\bm{Z}^{d+1}
	\label{eq.4}
\end{equation}

\textit{The weight adjustment of the acceleration module:}

\begin{equation} 
	d\bm{W}^{d+1,d}=\frac{1}{m}d\bm{Z}^{d+1}(\bm{A}^{d})^{T}
	\label{eq.5}
\end{equation}
\begin{equation} 
	\bm{W}^{d+1,d(new)}=\bm{W}^{d+1,d(old)}-\alpha d\bm{W}^{d+1,d}
	\label{eq.6}
\end{equation}
where $ d\bm{A}^{d+1} $ represents the error from the later module, $ m $ denotes the number of training samples in one batch,  $ \alpha $ is the learning rate, and the other symbols represent the intermediate variables or have been explained in Equation~\eqref{eq.1}.

\section{Experiments and Results} \label{sec3}

{Ref.}~\cite{liu2020dendrite}{ demonstrates that DD has lower time complexity than traditional polynomials or ML with nonlinear functions, which will speed up the computation of online systems. In addition, DD has white-box properties and controllable accuracy for nonlinear mappings. The main purpose of the following experiments is to demonstrate the feasibility of DD with AC to further speed up the computation while retaining the properties of DD.}

\subsection{Precision and Identification of Unary Nonlinear Mapping}

In order to investigate the precision and identification of unary nonlinear mapping, we considered the normalized Bessel function defined by:

\begin{equation} 
	f(x)=\frac{sin(x)}{x^{2}}-\frac{cos(x)}{x}
	\label{eq.10}
\end{equation}
where we defined  $ x \in [-10,0) \cup (0,10]$ , then $ x $ and $ f(x) $  were normalized to $ [-1,1] $ , respectively. 
{We gradually increase the order of DD with AC to approximate the normalized Bessel function (from order 4 to order 15). The architectures of DD are designed by Ref.}~\cite{liu2020dendrite} {and are shown in Table~\ref{Tablen2}. The architectures of DD with AC are obtained from {Algorithm~\ref{algorithm1}}, and the results are shown in Table ~}\ref{Tablen2}{. It is worth noting that the input dimension of DD or DD with AC is set to 2 and one of the inputs is always 1 (Bias).}

\begin{table}[H] 
	\caption{\textls[-21]{Architectures of DD and DD with AC for unary nonlinear mapping/four-input nonlinear~system.}\label{Tablen2}}
	\newcolumntype{C}{>{\centering\arraybackslash}X}
	\begin{tabularx}{\textwidth}{cccCC}
		\toprule
		\textbf{Order}	& \textbf{DD}	& \textbf{DD with AC} & \textbf{Number of Modules in~DD}& \textbf{Number of Modules in~DD~with AC}\\
		\midrule
		4 & 3DD & 1DD + AC2 & 3 & 2\\
		5 & 4DD & 1DD + AC3 & 4 & 2\\
		6 & 5DD & 2DD + AC3 & 5 & 3\\
		7 & 6DD & 2DD + AC4 & 6 & 3\\
		8 & 7DD & 3DD + AC4 & 7 & 4\\
		9 & 8DD & 3DD + AC5 & 8 & 4\\
		10 & 9DD & 4DD + AC5 & 9 & 5\\
		11 & 10DD & 4DD + AC6 & 10 & 5\\
		12 & 11DD & 5DD + AC6 & 11 & 6\\
		13 & 12DD & 5DD + AC7 & 12 & 6\\
		14 & 13DD & 6DD + AC7 & 13 & 7\\
		15 & 14DD & 6DD + AC8 & 14 & 7\\
		\bottomrule
\end{tabularx}
		\noindent{\footnotesize{A linear module follows each model, and the statistics for this module are not included in this table. ``k''DD + AC``j'': The model contains ``k'' DD modules and one acceleration module, and the power of $ \bm{X} $ in the acceleration module is ``j''.}}
	
\end{table}

\begin{figure}[H]
\includegraphics[width=0.88\columnwidth]{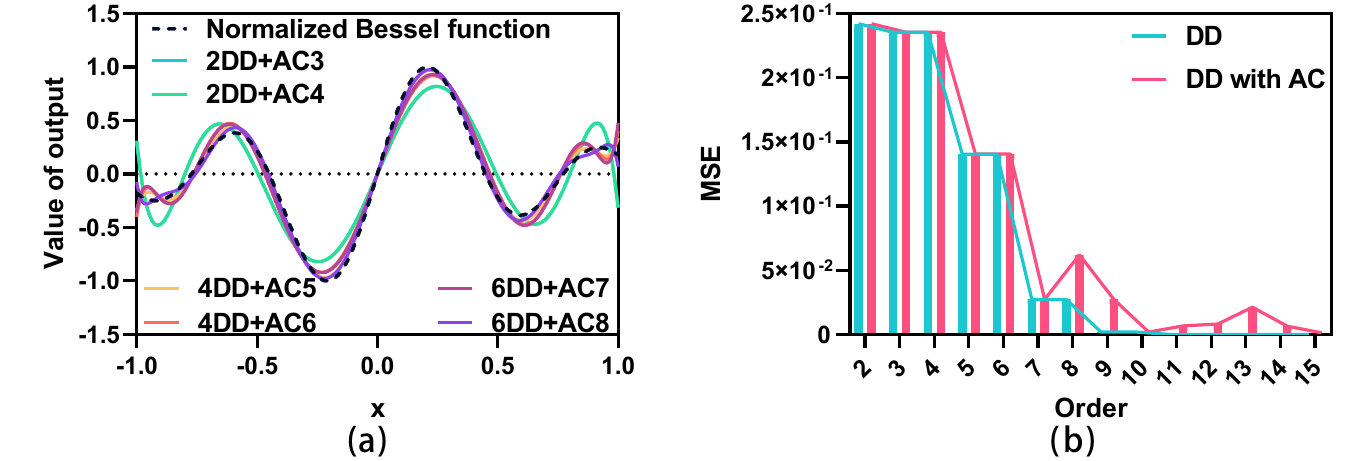}
	\caption{Precision  
 comparison between DD and DD with AC for nonlinear mapping. (\textbf{a}) Nonlinear mapping using DD with AC. (\textbf{b}) Precision comparison between DD and DD with AC as the target order increases. ``k''DD $ + $ AC``j'': 
		The model contains ``k'' DD modules and one acceleration module, and the power of $ \bm{X} $ in the acceleration module is ``j''.
	}
	\label{Fig3}
\end{figure}

Figure~\ref{Fig3} displays the precision of unary nonlinear mapping using DD with AC. What stands out in this figure is {the gradual increase in precision with the increasing order, which is present in both DD and DD with AC.} {Furthermore, the precision of DD with AC is lower than DD in the same order, especially at higher orders.}
{This may be due to the fact that} it is more difficult to find the optimal solution using DD with AC  than using DD. 
Although DD with AC has a drawback, it retains the key property of DD; that is, the precision increased with the number of modules, which corresponds to the property in Taylor's expansion ~\cite{liu2020dendrite}.

{The trained DD with AC models were} translated into the relation spectrum about inputs and outputs by formula simplification with MATLAB 2019b 
~\cite{ding2022multi,liu2020relation,Liu2021}.  Concretely, we set system input variables$ [1 x] $ and use it to express the forward propagation formula (see Figure~\ref{Fig2}). Then, the optimized weights were assigned to the corresponding matrices in DD with AC. Finally, the Relation spectrum was obtained through formula simplification in MATLAB. \textit{Note}
: The Relation spectrum, similar to the Fourier spectrum, focuses on transforming and observing the corresponding phenomenon in the spectrum for analysis. {The Relation spectrum presents the polynomial itself after the format transformation, so the transformation can be proved to be valid by observing whether there are similar relations in the Relation spectrum.} More explanations and applications can be found in previous researches, such as Ref.~\cite{Liu2021}, Ref.~\cite{liu2020relation}, and Ref.~\cite{ding2022multi}.

\begin{figure}[H]
	\hspace{-15pt} {\includegraphics[width=0.9\columnwidth]{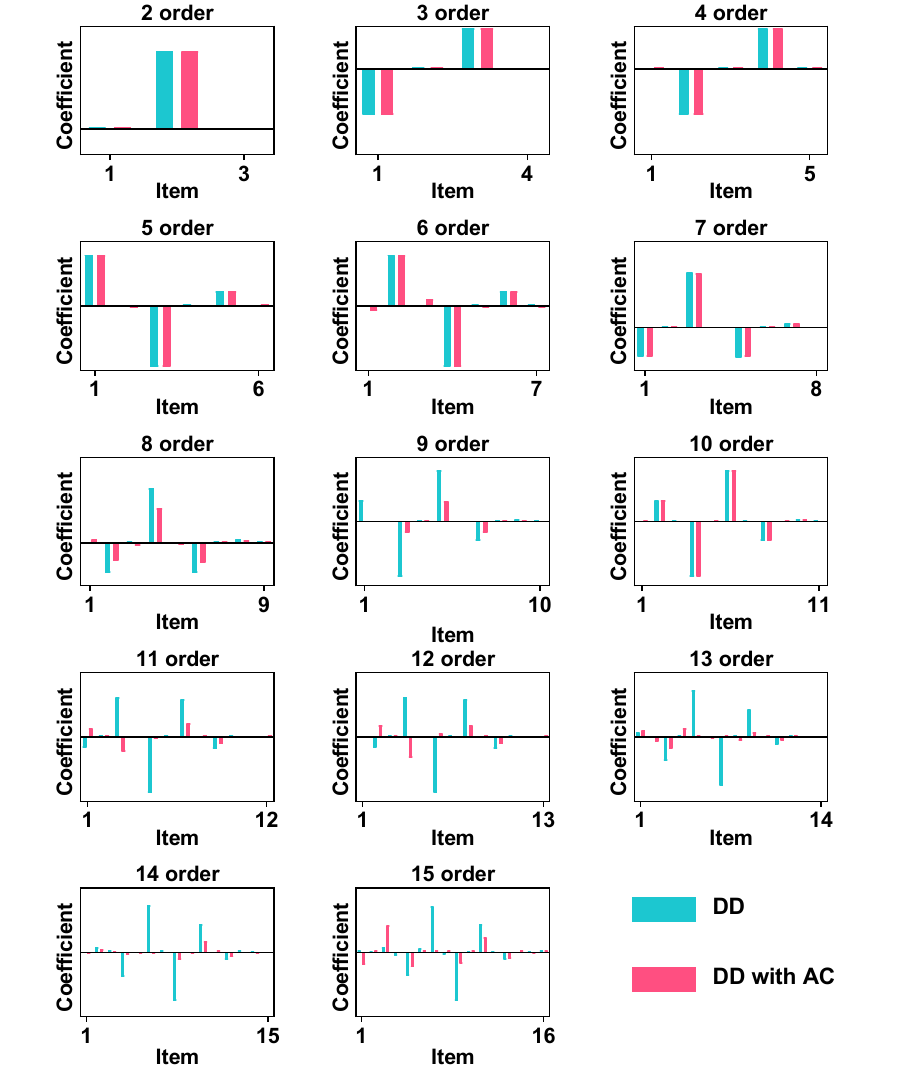}}
	\caption{Identification comparison between DD and DD with AC for nonlinear mapping. The abscissa indicates the relation items. In analogy to the abscissa of Fourier spectrum, there are many relation items. Therefore,  the relation items will not be listed here and can be obtained by looking up the table in the software.
	}
	\label{Fig4}
\end{figure}

\textls[-20]{Turning now to the identification by DD with AC in Figure~\ref{Fig4}, {the comparison between DD and DD with AC shows that DD with AC also retains the properties of the relation spectrum of DD.} The difference in the Relation spectrum corresponds to the difference in precision in Figure~\ref{Fig3}. Models with large differences in precision also have large differences in the Relation spectrum. In other words, the Relation spectrum in Figure~\ref{Fig4} explain the models in Figure~\ref{Fig3}. }

\subsection{Mapping Precision and Identification of Multi-Input Nonlinear System}

\begin{figure}[H]
	{\includegraphics[width=0.95\columnwidth]{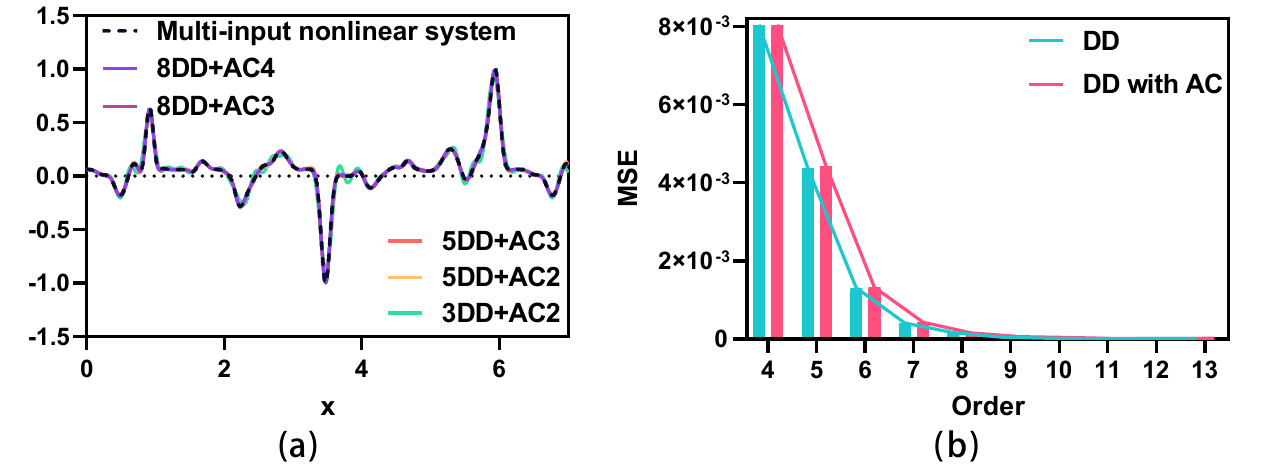}}
	\caption{Precision 
 comparison between DD and DD with AC for multi-input nonlinear system. (\textbf{a})~Multi-input nonlinear system using DD with AC. (\textbf{b}) Precision comparison between DD and DD with AC as the target order increases. ``k''DD + AC``j'': The model has ``k'' DD modules and one acceleration module, and the power of $ \bm{X} $ in the acceleration module is ``j''.
	}
	\label{Fig5}
\end{figure}

Modeling a multi-input nonlinear system is an essential and common demand in online systems, such as sensor systems. We randomly constructed a multi-input nonlinear system, whose output is shown in Figure~\ref{Fig5} (dotted line) and the inputs are defined by:
\begin{equation} 
	\left\{
	\begin{aligned}
		I_{1}(t)=&sin(2t) \\ 
		I_{2}(t)=&sin(3t)\\
		I_{3}(t)=&sin(5t)\\
	\end{aligned}
	\right.
	\label{eq.8}
\end{equation}
where we defined $ t \in [0,7] $. 

{ We gradually increase the order of DD and DD with AC to approximate the multi-input nonlinear system (from order 4 to order 13). The architectures of DD were designed according to the literature}~\cite{liu2020dendrite} (see Table~\ref{Tablen4}). { The architectures of DD with AC were obtained by {Algorithm~\ref{algorithm1}}, and the results are given in Table~}\ref{Tablen4}.{  It is worth noting that the input dimension of DD or DD with AC is set to 4, and one of the inputs is always 1 (Bias).}

\begin{table}[H] 
	\caption{ Architectures of DD and DD with AC for a four-input nonlinear system.\label{Tablen4}}
	\newcolumntype{C}{>{\centering\arraybackslash}X}
	\begin{tabularx}{\textwidth}{ccccC}
		\toprule
		\textbf{Order}	& \textbf{DD}	& \textbf{DD with AC} & \textbf{Number of Modules in DD}& \textbf{Number of Modules in~DD~with~AC}\\
		\midrule
		4 & 3DD & 2DD + AC1 & 3 & 3\\
		5 & 4DD & 2DD + AC2 & 4 & 3\\
		6 & 5DD & 3DD + AC2 & 5 & 4\\
		7 & 6DD & 4DD + AC2 & 6 & 5\\
		8 & 7DD & 5DD + AC2 & 7 & 6\\
		9 & 8DD & 5DD + AC3 & 8 & 6\\
		10 & 9DD & 6DD + AC3 & 9 & 7\\
		11 & 10DD & 7DD + AC3 & 10 & 8\\
		12 & 11DD & 8DD + AC3 & 11 & 9\\
		13 & 12DD & 8DD + AC4 & 12 & 9\\
		\bottomrule
\end{tabularx}
		\noindent{\footnotesize{A linear module follows each model, and the statistics for this module are not included in this table. ``k''DD + AC``j'': The model contains ``k'' DD modules and one acceleration module, and the power of $ \bm{X} $ in the acceleration module is ``j''.}}
	
\end{table}

Figure~\ref{Fig5} presents the precision of multi-input nonlinear system using DD with AC. What stands out in this figure is that the precision gradually improved with the increasing order, which exists in DD and DD with AC.

\begin{figure}[H] 
	\hspace{-20pt} {\includegraphics[width=0.95\columnwidth]{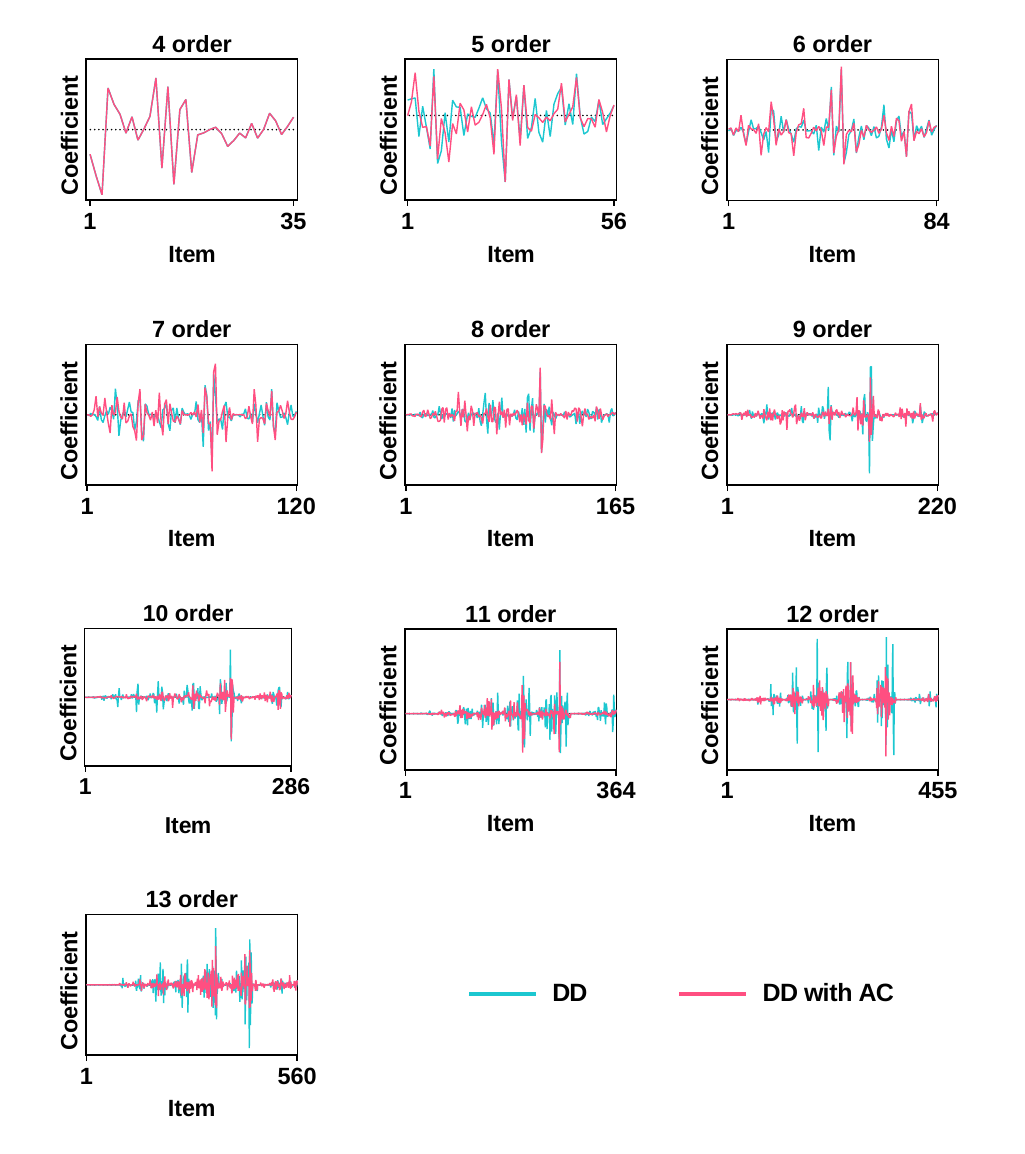}}
	\caption{Identification comparison between DD and DD with AC for multi-input nonlinear system. The abscissa indicates the relation items. In analogy to the abscissa of Fourier spectrum, there are many relation items. Therefore,  the relation items will not be listed here and can be obtained by looking up the table in the software.
	}
	\label{Fig6}
\end{figure}

The trained DD with AC models were transformed into a Relation spectrum of the inputs and outputs by simplifying the equations in MATLAB 2019b~\cite{ding2022multi,liu2020relation,Liu2021}. Figure~\ref{Fig6} provides an identification comparison between DD and DD with AC for a multi-input nonlinear system. The results between DD and DD with AC  are similar, revealing that DD with AC also retains the properties of the Relation spectrum in DD. These properties have a wide range of applications, such as analyzing the human brain~\cite{Liu2021} and physical design~\cite{ding2022multi}.

\subsection{Time Complexity}

\subsubsection{Computation of Time Complexity}

Here we take unary nonlinear mapping as an example to calculate the time complexity. The time complexity of the modules is summarized in Table~\ref{Table1}.

\begin{table}[H] 
	\caption{Time complexity in DD and DD with AC for unary nonlinear mapping.\label{Table1}}
	\newcolumntype{C}{>{\centering\arraybackslash}X}
	\begin{tabularx}{\textwidth}{CC}
		\toprule
		\textbf{Module}	& \textbf{Time Complexity}\\
		\midrule
		DD module (``$ \bm{W}\bm{A} \circ \bm{X} $") & $ 6 $ multiplication, $ 2 $ addition\\
		Linear module (``$ \bm{W}\bm{A}$") & $ 2 $ multiplication, $ 1 $ addition\\
		Acceleration module (``$ \bm{W}\bm{A} \circ \bm{X}^{c} $") & $ (4+2c) $ multiplication, $ 2 $ addition\\
		\bottomrule\end{tabularx}
		\noindent{\footnotesize{Unary Nonlinear Mapping means that the input vector $ \bm{X} $ contains two elements, one of which is 1, for example: $ \bm{X}=[1 \quad x] $. $ c $ denotes Power of Number.  $ \bm{X}^{c} $ represents $ \bm{X} $ to the power of $ c $. }}
	
\end{table}

DD is composed of DD modules and a linear module~\cite{liu2020dendrite}. Therefore, DD contains $ 6(n-1)+2 $ multiplication and $ 2(n-1)+1 $ addition, where $ n $ denotes the order of polynomial. The time complexity of DD is $ O(n) $, which happens to be consistent with Horner's method~\cite{cajori1911horner}.

\begin{figure}[H]
	{\includegraphics[width=0.95\columnwidth]{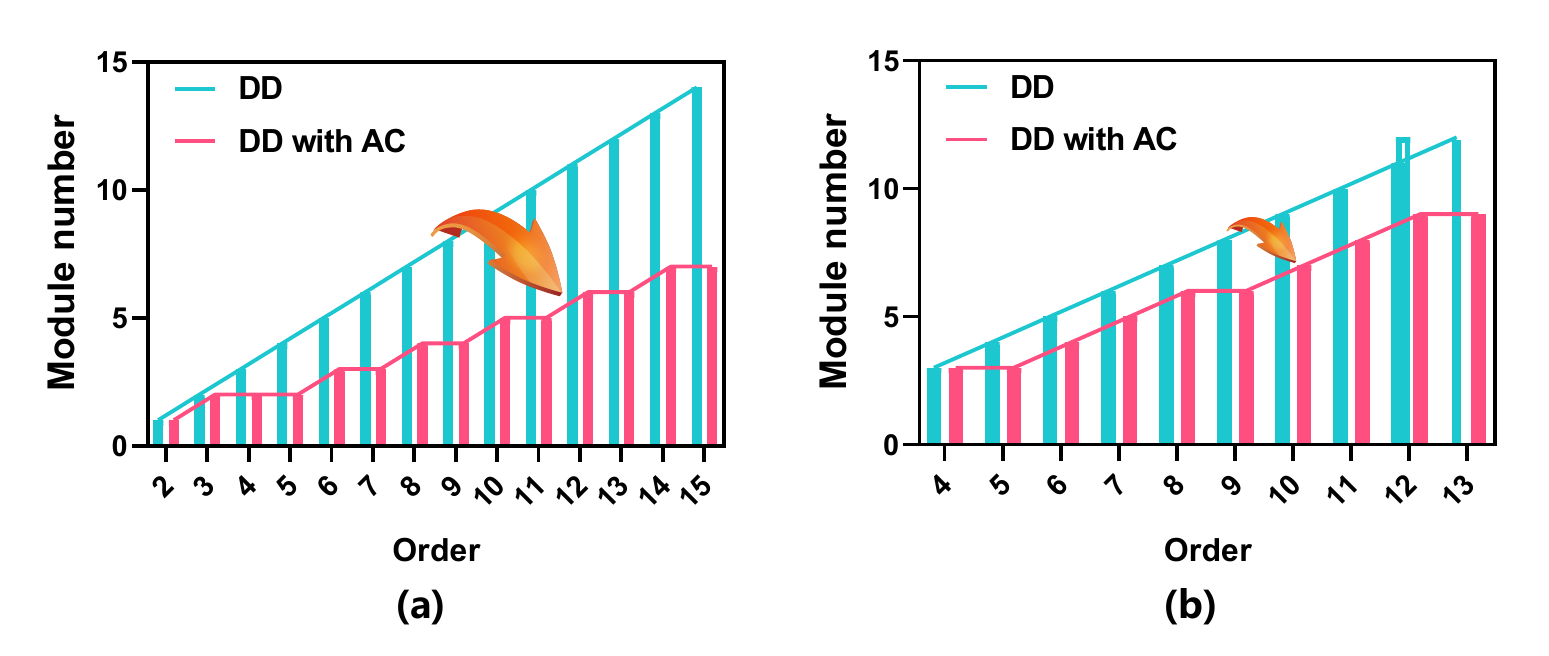}}
	\caption{The number of modules required for target order. (\textbf{a})~Module number for two-input system. When one of the inputs is 1, it corresponds to the unary nonlinear mapping above. (\textbf{b})~Module number for four-input system. When one of the inputs is 1, it corresponds to the above multi-input nonlinear system.
	}
	\label{Fig7}
\end{figure}

We take the two-input system (unary nonlinear mapping) and the four-input system using DD with AC as an example and display the number of modules required for the target order in Figure~\ref{Fig7}. Concretely, the architecture of DD with AC is obtained by {Algorithm~1}, and the results are shown in Tables~\ref{Tablen2} and \ref{Tablen4}. It is evident from Figure~\ref{Fig7} and Table~\ref{Table1} that, compared with increasing one order by adding a DD module, the time complexity is reduced by 2 multiplications and 1 addition by adjusting the acceleration module. \textbf{Therefore, the time complexity of DD with AC is less than that of DD or Horner's method~
\cite{cajori1911horner}.}

\subsubsection{Experiments of Time Complexity}

{In addition, to further verify the above results, 20 runs of DD and DD with AC were performed for the two-input system and the four-input system, and the run times (online speeds) were recorded.} The tests were executed in \textit{MATLAB 2021b} on a 2.2-GHz laptop Personal Computer (PC). Among them, 
{we recorded the running time for 10,000 forward-propagation with 1000 samples for the 2-input systems and  the running time for 10,000 forward-propagation with 7000 samples for the 4-input systems.}

\begin{figure}[H]
	{\includegraphics[width=0.95\columnwidth]{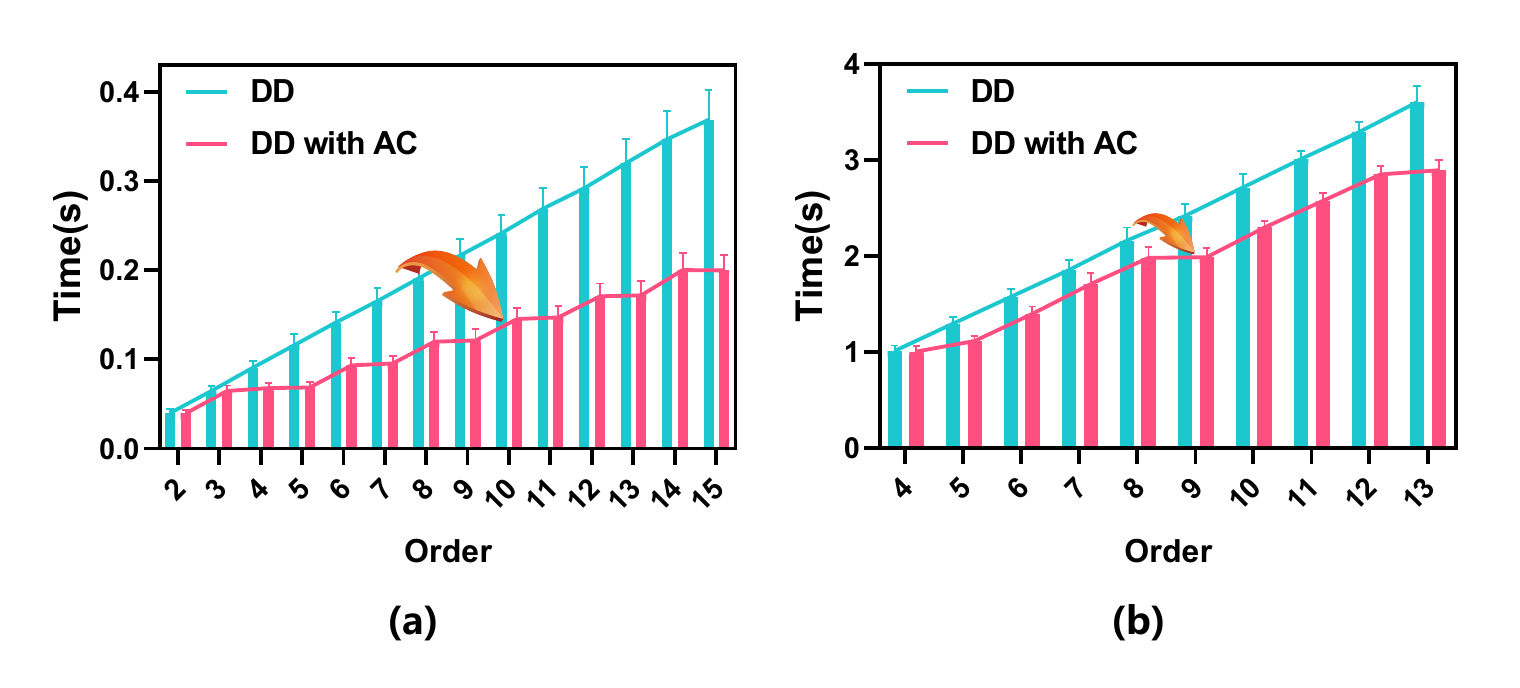}}
	\caption{Online speed for a certain order. (\textbf{a}) Online speed for a two-input system. (\textbf{b}) Online speed for a four-input system.	Data: $ Mean \pm  SD $.
	}
	\label{Fig8}
\end{figure}

The results are ideal. All online speeds from the online tests met our expectations (see Figure~\ref{Fig8}). By comparing the results of Figures~\ref{Fig7} and~\ref{Fig8}, it can be concluded that the online speed  
{ corresponds with} the number of modules.

\section{Discussion} \label{sec4}
	
Prior studies have noted the importance of nonlinear mapping in online computation~\cite{5306098,5560598,sammon1969nonlinear,DCOM11,penJDK}. Our previous studies about DD observed faster speeds of nonlinear mapping using DD~\cite{liu2020dendrite}. 
Due to the white-box attribute, controllable precision, and lower time complexity, DD was comprehensively applied in the areas of energy, traffic, weather, and physical design~\cite{Han2022,wang2022hybrid,zhang2022adaptive,li2021semantic,ding2022multi,Jin2022,li2022algorithm}. However, there has still been a redundant calculation in DD. This study aimed to eliminate the redundant computation while retaining DD's properties. 

According to the analysis of DD, we designed an acceleration module, presented DD with AC, and conducted three experiments. These experimental results are in accord with the theoretical results. DD with AC has a lower time complexity than DD (see \mbox{Figures~\ref{Fig7} and \ref{Fig8}}), has controllable precision (see Figures~\ref{Fig3} and \ref{Fig5}), and can also  be used for nonlinear mapping and system identification (see Figures~\ref{Fig4} and \ref{Fig6}). This paper was the continued work of previous studies about DD, which may improve the application of DD in various fields. 
{Table~\ref{TableDD} shows some examples of current applications of DD by referring to the reviewer's suggestions.}

{The limitation of this paper comes from the lack of engineering data. In this paper, DD with AC is fundamentally verified by many experiments, which is a stronger way to prove it than using special data. Future experiments will be conducted on engineering problems.}
\begin{table}[H] 
	\caption{{Some applications of DD.\label{TableDD}}}
	\newcolumntype{C}{>{\centering\arraybackslash}X}
	\begin{tabularx}{\textwidth}{p{11cm}C}
		\toprule
		\textbf{Applications}	& \textbf{Literature}\\
		\midrule
		A hybrid data-driven framework for spatiotemporal traffic flow data imputation & Literature~\cite{wang2022hybrid}\\
		Energy saving of buildings for reducing carbon dioxide emissions using novel dendrite net integrated adaptive mean square gradient & Literature~\cite{Han2022}\\
		Unsteady aerodynamics modeling method based on dendrite-based gated recurrent neural network model & Literature~\cite{liu2022unteady}\\
		A radial sampling-based subregion partition method for dendrite network-based reliability analysis & 
		Literature~\cite{lu2022radial}\\
		An Algorithm for Precipitation Correction in Flood Season Based on Dendritic Neural Network & Literature~\cite{li2022algorithm} \\
		Multi-Objective Optimization for the Radial Bending and Twisting Law of Axial Fan Blades & Literature~\cite{ding2022multi} \\
		An Accuracy Prediction Method of the RV Reducer to Be Assembled Considering Dendritic Weighting Function & Literature~\cite{Jin2022} \\
		An Adaptive Dendrite-HDMR Metamodeling Technique for High-Dimensional Problems & Literature~\cite{zhang2022adaptive}\\
		Convolutional dendrite net detects myocardial infarction based on ECG signal measured by flexible sensor & Literature~\cite{ma2021convolutional} \\
		Photovoltaic Power Prediction Under Insufficient Historical Data Based on Dendrite Network and Coupled Information Analysis & Literature~\cite{lu4184484photovoltaic} \\
		\bottomrule
	\end{tabularx}
\end{table}

\section{Conclusions} \label{sec5}

This paper presents a Dendrite Net with an Acceleration module for nonlinear mapping and system identification.
The theoretical and experimental results suggest that DD with AC retains DD's nonlinear mapping properties and system identification properties. Interestingly, the time complexity of DD is lower than the traditional polynomial or ML with a nonlinear function and is consistent with Horner's method. The time complexity of DD with AC is lower than DD or Horner's method, which provides a new strategy for online systems that require lower time complexity and has the potential to speed up the calculation of basic functions in computers.

\vspace{6pt} 



\authorcontributions{Conceptualization, G.L.; validation, G.L., Y.P. and S.Y.; formal analysis, X.N.;   writing---original draft preparation, G.L and Y.P.; writing---review and editing, S.Y. and X.N.;  supervision, J.W. and H.W.; project administration, H.W.; funding acquisition, X.N., J.W. and H.W. All authors have read and agreed to the published version of the manuscript.}

\begin{adjustwidth}{-\extralength}{0cm}

\reftitle{References}

\end{adjustwidth}
\end{document}